%% file: main.tex
\documentclass[runningheads]{llncs}
\usepackage[english]{babel}
\usepackage{makecell}
\usepackage{graphicx}
\usepackage{amssymb}
\usepackage{amsmath}
\usepackage{multirow}
\usepackage{rotating}
\usepackage{hyperref}
\usepackage{cite}
\usepackage{mymacros}
\usepackage{tikz}

\newcolumntype{Y}{>{\centering\arraybackslash}X}
\newcolumntype{s}{>{\hsize=.5\hsize}X}

\addto\extrasenglish{%
}

\begin{document}
\title{
OPerA: Object-Centric Performance Analysis
}
%
%
\author{
Gyunam Park\orcidID{0000-0001-9394-6513} 
\and
Jan Niklas Adams\orcidID{0000-0001-8954-4925} \and
Wil. M. P. van der Aalst\orcidID{0000-0002-0955-6940}
}
\authorrunning{G. Park, J. N. Adams, and W.M.P. van der Aalst}
%
\institute{Process and Data Science Group (PADS), RWTH Aachen University \\ \email{\{gnpark,niklas.adams,wvdaalst\}@pads.rwth-aachen.de}}
\maketitle              
\begin{abstract}
Performance analysis in process mining aims to provide insights on the performance of a business process by using a process model as a formal representation of the process.
Such insights are reliably interpreted by process analysts in the context of a model with formal semantics.
Existing techniques for performance analysis assume that a single case notion exists in a business process (e.g., a patient in healthcare process).
However, in reality, different objects might interact (e.g., order, item, delivery, and invoice in an O2C process). 
In such a setting, traditional techniques may yield misleading or even incorrect insights on performance metrics such as waiting time.
More importantly, by considering the interaction between objects, we can define object-centric performance metrics such as synchronization time, pooling time, and lagging time.
In this work, we propose a novel approach to performance analysis considering multiple case notions by using object-centric Petri nets as formal representations of business processes.
The proposed approach correctly computes existing performance metrics, while supporting the derivation of newly-introduced object-centric performance metrics.
We have implemented the approach as a web application and conducted a case study based on a real-life loan application process.

\end{abstract}

\input{Sections/1-Intro}

\input{Sections/2-Related}
\input{Sections/3-Background}
\input{Sections/4-Performance-Metric}

\input{Sections/5-Evaluation}
\input{Sections/6-Conclusion}

\bibliographystyle{splncs04}
\bibliography{mybib}

\end{document}

%% file: Sections/1-Intro.tex
\section{Introduction}
Process mining provides techniques to extract insights from event data recorded by information systems, including process discovery, conformance checking, and performance analysis~\cite{DBLP:books/sp/Aalst16}.
Especially performance analysis provides techniques to analyze the performance of a business process, e.g., bottlenecks, using process models as representations of the process~\cite{067cb84c5d19493cb0f364ed65748cec}.

\begin{figure}[!htb]
    \centering
    \includegraphics[width=1\linewidth]{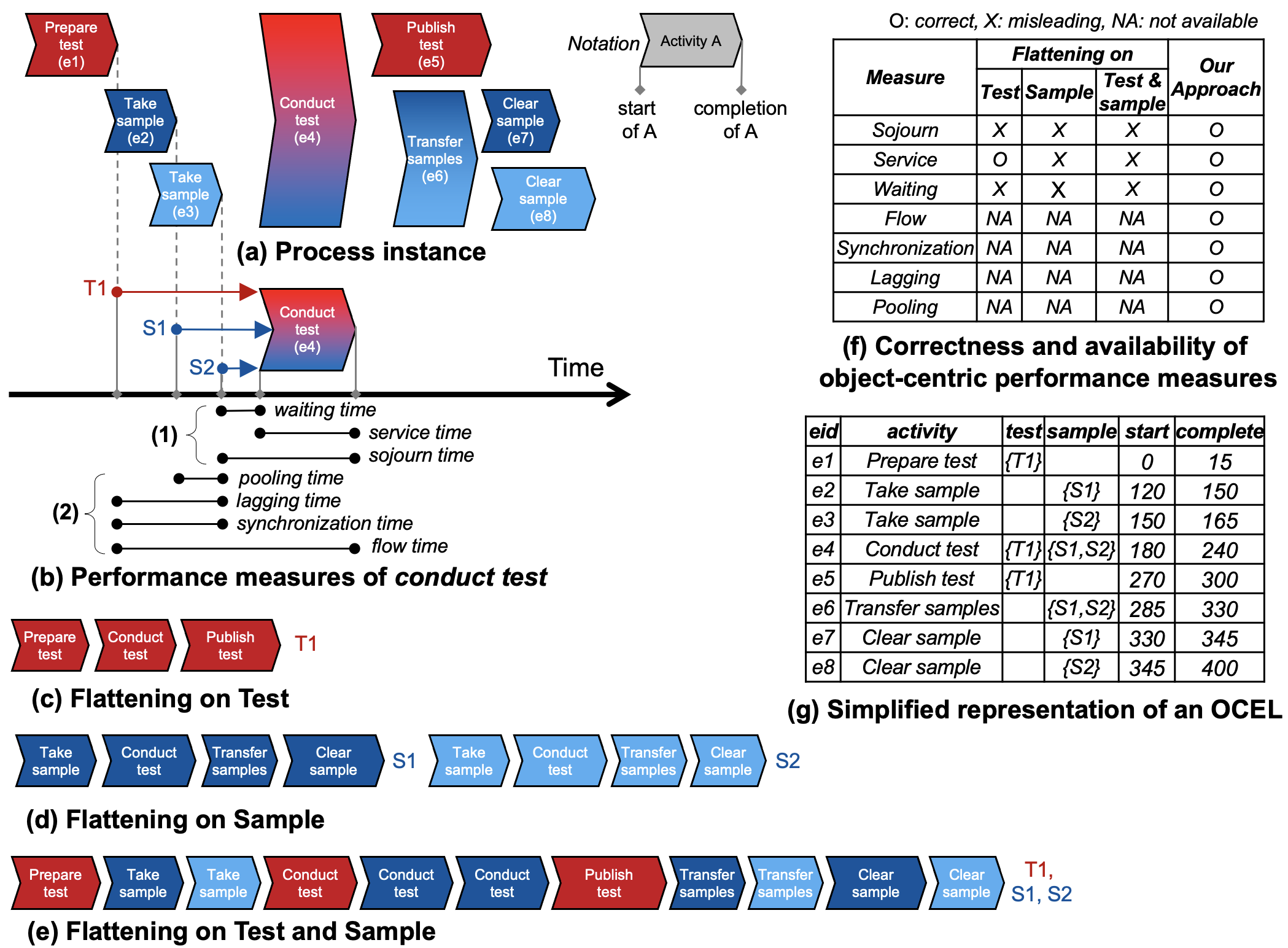}
    \caption{A motivating example showing misleading insights from existing approaches to performance analysis and the proposed object-centric performance analysis}
    \label{fig:motivation}
\end{figure}

Existing techniques for performance analysis have been developed, assuming that a single case notion exists in business processes, e.g., a patient in a healthcare process~\cite{10.1007/978-3-642-34002-4_22, weske_unbiased_2018, peter, 10.1007/978-3-642-45005-1_27, DBLP:conf/icpm/LeemansPW19, 067cb84c5d19493cb0f364ed65748cec, 6eac4d541d66416180c36e95696ea87f}.
Such a case notion correlates events of a process instance and represents them as a single sequence, e.g., a sequence of events of a patient.
However, in real-life business processes supported by ERP systems such as SAP and Oracle, multiple objects (i.e., multiple sequences of events) exist in a process instance~\cite{DBLP:conf/er/BayomieCRM19, DBLP:journals/fuin/AalstB20} and they share events (i.e., sequences are overlapping).
\autoref{fig:motivation}(a) shows a process instance in a simple blood test process as multiple overlapping sequences.
The red sequence represents the event sequence of test \textit{T1}, whereas the blue sequences indicate the event sequences of samples \textit{S1} and \textit{S2}, respectively.
The objects share \textit{conduct test} event (\textit{e4}), i.e., all the sequences overlap, and the samples share \textit{transfer samples} event (\textit{e6}), i.e., the sample sequences overlap.

The goal of object-centric performance analysis is to analyze performance in such ``object-centric'' processes with multiple overlapping sequences using 1) existing performance measures and 2) new performance measures considering the interaction between objects.
\autoref{fig:motivation}(b)(1) visualizes existing performance measures related to event \textit{conduct test}.
\textit{Waiting time} of \textit{conduct test} is the time spent before conducting the test after preparing test \textit{T1} and samples \textit{S1} and \textit{S2}, while the \textit{service time} is the time spent for conducting the test and \textit{sojourn time} is the sum of \textit{waiting time} and \textit{service time}.
Furthermore, \autoref{fig:motivation}(b)(2) shows new performance measures considering the interaction between objects.
First, \textit{synchronization time} is the time spent for synchronizing different objects, i.e., samples \textit{S1} and \textit{S2} with test \textit{T1} to conduct the test.
Next, \textit{pooling time} is the time spent for pooling all objects of an object type, e.g., the pooling time of \textit{conduct test} w.r.t. \textit{sample} is the time taken to pool the second sample.
Third, \emph{lagging} time is the time spent due to the lag of an object type, e.g., the lagging time of \textit{conduct test} w.r.t. \textit{test} is the time taken due to the lag of the second sample.
Finally, \textit{flow time} is the sum of \textit{sojourn time} and \textit{synchronization time}.

A natural way to apply existing techniques to multiple overlapping sequences is to \emph{flatten} them into a single sequence.
To this end, we select an object type(s) as a case notion, removing events not having the object type and replicating events with multiple objects of the selected type~\cite{DBLP:journals/fuin/AalstB20}.
For instance, \autoref{fig:motivation}(a) is flattened to \autoref{fig:motivation}(c) by using test as a case notion, to \autoref{fig:motivation}(d) by using sample as a case notion, and \autoref{fig:motivation}(e) by using both test and sample as a case notion.

However, depending on the selection, flattening results in misleading insights.
\autoref{fig:motivation}(f) summarizes the correctness of object-centric performance analysis on flattened sequences.
1) Flattening on test provides a misleading waiting time, measured as the time difference between the complete time of \textit{prepare test} and the start time of \textit{conduct test}, and, thus, a misleading sojourn time.
2) Flattening on sample results in misleading insights on the service time since two service times are measured despite the single occurrence of the event.
3) By flattening on both test and sample, the waiting time for \textit{take sample} is measured in relation to \textit{prepare test} although they are independent events from different object types.

In this work, we suggest a novel approach to object-centric performance analysis. 
The approach uses an Object-Centric Event Log (OCEL) that store multiple overlapping sequences without flattening (cf. \autoref{fig:motivation}(g)) as an input.
Moreover, we use Object-Centric Petri Nets (OCPNs)~\cite{DBLP:journals/fuin/AalstB20} as a formalism to represent process models, and the object-centric performance is analyzed in the context of process models.
With formal semantics of OCPNs, we can reliably compute and interpret performance analysis results, considering the concurrency, loops, etc~\cite{DBLP:journals/widm/AalstAD12}.

More in detail, we first discover an OCPN that formally represents a process model from the OCEL.
Next, we replay the OCEL on the discovered OCPN to produce \emph{token visits} and \emph{event occurrences}.
Finally, we compute object-centric performance measures using the token visit and event occurrence.
For instance, in the proposed approach, the waiting time of \textit{Conduct test} is computed as the difference between \textit{e4}'s start and \textit{e1}'s complete. 
The synchronization time is computed as the time difference between \textit{e3}'s complete and \textit{e1}'s complete.

In summary, we provide the following contributions.
\begin{enumerate}
    \item Our approach correctly calculates existing performance measures in an object-centric setting. 
    \item Our approach supports novel object-centric performance metrics taking the interaction between objects into account, such as synchronization time.
    \item The proposed approach has been implemented as a web application\footnote{\label{opera} A demo video, sources, and manuals are available at \url{https://github.com/gyunamister/OPerA}} and a case study with a real-life event log has been conducted to evaluate the effectiveness of the approach.
\end{enumerate}

The remainder is organized as follows. We discuss the related work in \autoref{sec:related}. Next, we present the preliminaries, including OCELs and OCPNs in \autoref{sec:preliminaries}. 
In \autoref{sec:approach}, we explains the approach to object-centric performance analysis.
Afterward, \autoref{sec:evaluation} introduces the implementation of the proposed approach and a case study using real-life event data.
Finally, \autoref{sec:conclusion} concludes the paper.

%% file: Sections/2-Related.tex
\section{Related Work}\label{sec:related}
\subsection{Performance Analysis in Process Mining}
Performance analysis has been widely studied in the context of process mining. 
\autoref{tab:comparison} compares existing work and our proposed work in different criteria: 1) if formal semantics exist to analyze performance in the context of process models, 2) if aggregated measures, e.g., mean and median, are supported, 3) if frequency analysis is covered, 4) if time analysis is covered, and 5) if multiple case notions are allowed to consider the interactions of different objects.
Existing algorithms/techniques assume a single case notion, not considering the interaction among different objects.

\begin{table}[h]
\centering
\caption{Comparison of algorithms/techniques for performance analysis\label{tab:comparison}}
\begin{tabular}{|l|l|ccccc|}
\hline
\textbf{Author}      & \textbf{Technique}                 & \textbf{Form.} & \textbf{Agg.} & \textbf{Freq.} & \textbf{Perf.} & \textbf{Obj.} \\ \hline
Mat\'e et al.~\cite{10.1007/978-3-642-34002-4_22} & \textit{Business Strategy Model} & - & \checkmark & \checkmark & \checkmark &  - \\
Denisov et al.~\cite{weske_unbiased_2018} & \textit{Performance Spectrum} & - & \checkmark & \checkmark & \checkmark &  -\\ 
Hornix~\cite{peter} & \textit{Petri Nets} & \checkmark & \checkmark & \checkmark & \checkmark &  - \\
Rogge-Solti et al.~\cite{10.1007/978-3-642-45005-1_27} & \textit{Stochastic Petri Nets} & \checkmark & \checkmark & - & \checkmark &  - \\
Leemans et al.~\cite{DBLP:conf/icpm/LeemansPW19} & \textit{Directly Follows Model} & \checkmark & \checkmark & \checkmark & \checkmark &  - \\
Adriansyah et al.~\cite{067cb84c5d19493cb0f364ed65748cec} & \textit{Robust Performance} & \checkmark & \checkmark & \checkmark & \checkmark &  - \\
Adriansyah~\cite{6eac4d541d66416180c36e95696ea87f} & \textit{Alignments} & \checkmark & \checkmark & \checkmark & \checkmark &  - \\
\hline
\textbf{Our work} & \textit{\textbf{Object-Centric}} & \textbf{\checkmark} & \textbf{\checkmark} & \textbf{\checkmark} & \textbf{\checkmark} & \textbf{\checkmark} \\ \hline
\end{tabular}
\end{table}

\subsection{Object-Centric Process Mining}
Traditionally, methods in process mining have the assumption that each event is associated with exactly one case, viewing the event log as a set of isolated event sequences. Object-centric process mining breaks with this assumption, allowing one event to be associated with multiple cases and, thus, having shared events between event sequences. An event log format has been proposed to store object-centric event logs~\cite{ocel}, as well as a discovery technique for OCPNs~\cite{DBLP:journals/fuin/AalstB20} and a conformance checking technique to determine precision and fitness of the net~\cite{DBLP:conf/icpm/AdamsA21}.
Furthermore, Esser and Fahland~\cite{GraphDatabases} propose a graph database as a storage format for object-centric event data, enabling a user to use queries to calculate different statistics.
A study on performance analysis is, so far, missing in the literature, with only limited metrics being supported in~\cite{DBLP:journals/fuin/AalstB20} by flattening event logs and replaying them. However, object-centric performance metrics are needed to accurately assess performance in processes where multiple case notions occur.


The literature contains several notable approaches to deal with multiple case notions. Proclets ~\cite{ProcletManytoMany} is the first introduced modeling technique to describe interacting workflow processes and, later, artifact-centric modeling~\cite{ArtifactsIntroduction} extends this approach. DB nets~\cite{DBNetsCPNsAndRelationalDatabase} are a modeling technique based on colored Petri nets. OCBC~\cite{OCBCFoundation} is a newly proposed technique that includes the evolution of a database into an event log, allowing for the tracking of multiple objects. Object-centric process mining aims to alleviate the weaknesses of these techniques. The approaches and their weaknesses are more deeply discussed in~\cite{DBLP:journals/fuin/AalstB20}. 

%% file: Sections/3-Background.tex
\section{Background}\label{sec:preliminaries}
\subsection{Object-Centric Event Data}
\begin{definition}[Universes]\label{def:universes}
Let $\univ{ei}$ be the universe of event identifiers, 
$\univ{act}$ the universe of activity names, 
$\univ{time}$ the universe of timestamps, 
$\univ{ot}$ the universe of object types, and
$\univ{oi}$ the universe of object identifiers.
$\mathit{type} \in \univ{oi} \rightarrow \univ{ot}$ assigns precisely one type to each object identifier.
$\univ{omap}{=}\{ \mathit{omap} \in \univ{ot}  \not\rightarrow \pow(\univ{oi}) \mid \forall_{\mathit{ot}\in \mathit{dom}(\mathit{omap})}\ \forall_{\mathit{oi}\in \mathit{omap}(\mathit{ot})}\ \mathit{type}(\mathit{oi}){=}\mathit{ot} \}$ is the universe of all object mappings indicating which object identifiers are included per type.
$\univ{event}{=}\univ{ei} \times \univ{act} \times \univ{time} \times \univ{time} \times \univ{omap}$ is the universe of events.
\end{definition}

Given $e{=}(ei,act,st,ct,omap) \in \univ{event}$, $\pi_{ei}(e){=}ei$, $\pi_{act}(e){=}act$, $\pi_{st}(e){=}st$, $\pi_{ct}(e){=}ct$, and $\pi_{omap}(e){=}omap$.
Note that we assume an event has start and complete timestamps.

\autoref{fig:motivation}(b) describes a fraction of a simple object-centric event log with two types of objects.
For the event in the fourth row, denoted as $e4$, $\pi_{ei}(e4){=}\textit{e4}$, $\pi_{act}(e4){=}\textit{conduct test}$, $\pi_{st}(e4)=\textit{180}$, $\pi_{ct}(e4){=}\textit{240}$, $\pi_{omap}(e4)(test){=}\{\textit{T1} \}$, and $\pi_{omap}(e4)(sample){=}\{\textit{S1},\textit{S2} \}$.
Note that the timestamp in the example is simplified using the relative scale.

\begin{definition}[Object-Centric Event Log (OCEL)]\label{def:event}
An object-centric event log is a tuple $L{=}(E,\prec_E)$, where $E \subseteq \univ{event}$ is a set of events and $\prec_E \subseteq E \times E$ is a total order underlying $E$.
$\univ{L}$ is the set of all possible object-centric event logs.
\end{definition}

\subsection{Object-Centric Petri Nets}
A Petri net is a directed graph having places and transitions as nodes and flow relations as edges. 
A labeled Petri net is a Petri net where the transitions can be labeled.

\begin{definition}[Labeled Petri Net]\label{def:lpn}
A labeled Petri net is a tuple $N {=} (P,T,F,l)$ with $P$ the set of places, $T$ the set of transitions,
$P \cap T {=} \emptyset$, $F\subseteq (P \times T) \cup (T \times P)$ the flow relation, and $l \in T \not\rightarrow \univ{act}$  a labeling function.
\end{definition}


Each place in an OCPN is associated with an object type to represent interactions among different object types. 
Besides, variable arcs represent the consumption/production of a variable amount of tokens in one step.

\begin{definition}[Object-Centric Petri Net]\label{def:oopn}
An \emph{object-centric Petri net} is a tuple $\mathit{ON} {=} (N,\mathit{pt},F_{\mathit{var}})$ where
$N {=} (P,T,F,l)$ is a labeled Petri net, $\mathit{pt} \in P \rightarrow \univ{ot}$ maps places to object types, and $F_{\mathit{var}}\subseteq F$ is the subset of variable arcs.
\end{definition}

\autoref{fig:replay}(a) depicts an OCPN, $ON_1{=}(N,pt,F_{var})$ with $N{=}(P,T,F,l)$ where $P {=} \{ p1, \dots, p9 \}$, $T {=} \{ t1,\dots,t6\}$, $F {=} \{(p1,t1), (p2,t2),\dots \}$, $l(t1){=} \textit{prepare test}$, etc., $pt(p1) {=} test$, $pt(p2){=}\textit{sample}$, etc., and $F_{var} {=} \{(p4,t3),(t3,p6),\dots \}$.


\begin{definition}[Marking]\label{def:oomark}
Let $\mathit{ON}{=}(N,\mathit{pt},F_{\mathit{var}})$ be an object-centric Petri net, where $N{=}(P,T,F,l)$.
$Q_{\mathit{ON}}{=}\{(p,\mathit{oi}) \in P \times \univ{oi} \mid type(oi) {=} \mathit{pt}(p)\}$ is the set of possible tokens.
A marking $M$ of $\mathit{ON}$ is a multiset of tokens, i.e., $M \in \bag(Q_{\mathit{ON}})$.
\end{definition}

For instance, marking $M_1{=}[(p3,T1),(p4,S1),(p4,S2)]$ denotes three tokens, among which place \textit{p3} has one token referring to object \textit{T1} and \textit{p4} has two tokens referring to objects \textit{S1} and \textit{S2}.

A binding describes the execution of a transition consuming objects from its input places and producing objects for its output places.
A binding $(t,b)$ is a tuple of transition $t$ and function $b$ mapping the object types of the surrounding places to sets of object identifiers. 
For instance, $(t3,b1)$ describes the execution of transition $t3$ with $b1$ where $b1(test){=}\{T1\}$ and $b1(sample){=}\{S1,S2\}$, where \textit{test} and \textit{sample} are the object types of its surrounding places (i.e., $p3,p4,p5,p6$).

A binding $(t,b)$ is \textit{enabled} in marking $M$ if all the objects specified by $b$ exist in the input places of $t$. For instance, $(t3,b1)$ is enabled in marking $M_1$ since $T1$, $S1$, and $S2$ exist in its input places, i.e., $p3$ and $p4$.

A new marking $M'$ is reached by executing enabled binding $(t,b)$ at $M$ leads to, denoted by $M \stackrel{(t,b)}{\longrightarrow} M'$.
As a result of executing $(t1,b1)$, $T1$ is removed from $p3$ and added to $p5$. Besides, $S1$ and $S2$ are removed from $p4$ and added to $p6$, resulting in new marking $M'{=}[(p5,T1),(p6,S1),(p6,S2)]$.

%% file: Sections/4-Performance-Metric.tex
\section{Object-Centric Performance Analysis}\label{sec:approach}
\begin{figure}[!htb]
    \centering
    \includegraphics[width=0.9\linewidth]{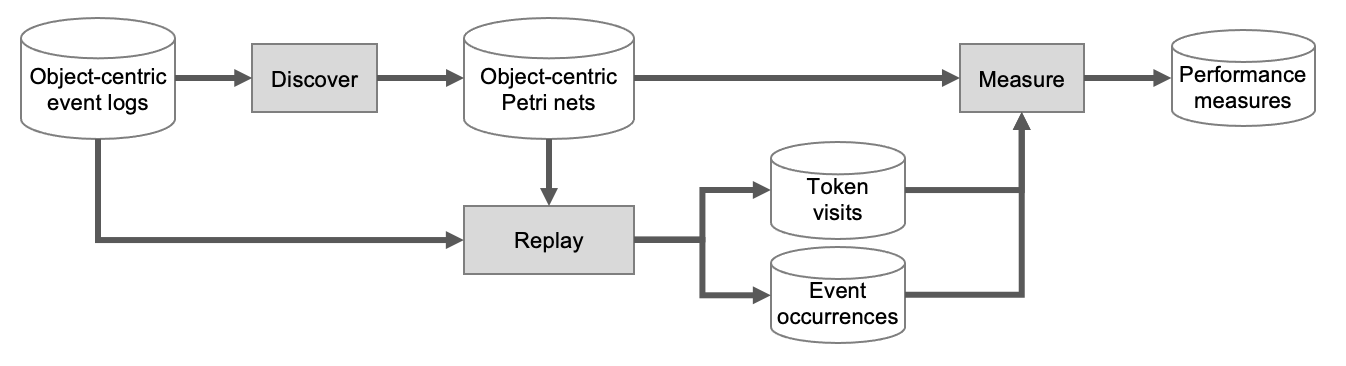}
    \caption{An overview of the proposed approach.}
    \label{fig:approach}
\end{figure}

This section introduces an approach to object-centric performance analysis.
\autoref{fig:approach} shows an overview of the proposed approach.
First, we discover an OCPN based on an OCEL.
Next, we replay the OCEL with timestamps on the discovered OCPN to connect events in the OCEL to the elements of OCPN and compute \emph{event occurrences} and \emph{token visits}.
Finally, we measure various object-centric performance metrics based on the event occurrence and token visit.
The discovery follows the general approach presented in~\cite{DBLP:journals/fuin/AalstB20}.
In the following subsections, we focus on explaining the rest.

\subsection{Replaying OCELs on OCPNs}
First, we couple events in an OCEL to an OCPN by ``playing the token game'' using the formal semantics of OCPNs.
Note that most of business processes are not sequential, and, thus, simply relating an event to its directly following event does not work.
By using the semantics of OCPNs, we can reliably relate events to process models by considering the concurrency and loop and correctly identify relationships between events.

As a result of the replay, a collection of \emph{event occurrences} are annotated to each visible transition, and a collection of \emph{token visits} are recorded for each place. 
First, an event occurrence represents the occurrence of an event in relation to a transition.

\begin{definition}[Event Occurrence]
Let $\mi{ON}{=}(N,\mi{pt},F_{\mi{var}})$ be an object-centric Petri net, where $N{=}(P,T,F,l)$.
An event occurrence $eo \in T \times \univ{event}$ is a tuple of a transition and an event.
$O_{\mi{ON}}$ is the set of possible event occurrences of $\mi{ON}$.
\end{definition}

\begin{figure}[!htb]
    \centering
    \includegraphics[width=1\linewidth]{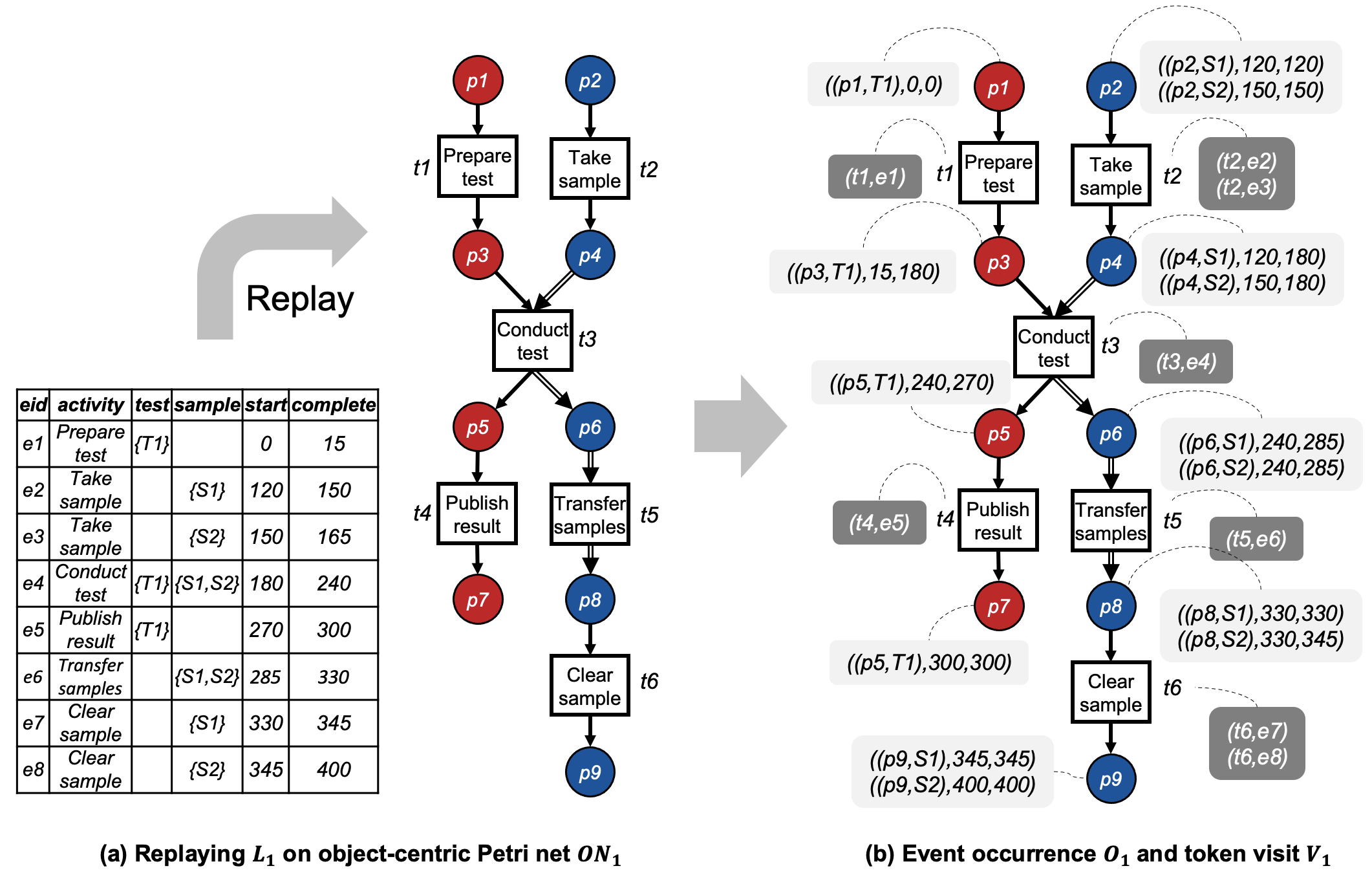}
    \caption{An example of replaying object-centric event logs on an object-centric Petri net}
    \label{fig:replay}
\end{figure}

For instance, $(\mi{t3},\mi{e4}) \in O_{\mi{ON_1}}$ is a possible event occurrence in $\mi{ON}_1$ shown in \autoref{fig:replay}(a).
It indicates that \textit{t3} is associated with the occurrence of event $\mi{e4}$.
In other words, \textit{t3} is fired by 1) consuming tokens $(\mi{p3},\mi{T1})$ from \textit{p3} and $(\mi{p4},\mi{S1})$ and $(\mi{p4},\mi{S2})$ from\textit{p4} at $180$ and 2) producing tokens $(\mi{p5},\mi{T1})$ to \textit{p5} and $(\mi{p6},\mi{S1})$ and $(\mi{p6},\mi{S2})$ \textit{p6} at $240$.
Note that we derive the consumed and produced tokens by using the transition and the event, i.e., we are aware of the input and output places of the transition and the involved objects of the event. 
Moreover, we know when the event starts and completes.

A token visit describes ``visit'' of a token to the corresponding place with the begin time of the visit, i.e., the timestamp when the token is produced, and the end time of the visit, i.e., the timestamp when the token is consumed.

\begin{definition}[Token Visit]
Let $\mi{ON}{=}(N,\mi{pt},F_{\mi{var}})$ be an object-centric Petri net, where $N{=}(P,T,F,l)$.
$Q_{\mathit{ON}}{=}\{(p,\mathit{oi}) \in P \times \univ{oi} \mid type(oi) {=} \mathit{pt}(p)\}$ is the set of possible tokens.
A token visit $\mi{tv} \in Q_{\mi{ON}} \times \univ{time} \times \univ{time}$ is a tuple of a token, a begin time, and an end time.
$\mi{TV}_{\mi{ON}}$ is the set of possible token visits of $\mi{ON}$.
\end{definition}

Given token visit $tv{=}((p,oi),bt,et)$, $\pi_{p}(tv){=}p$, $\pi_{oi}(tv){=}oi$, $\pi_{bt}(tv){=}bt$, and $\pi_{et}(tv){=}et$.
For instance, $((\mi{p3},\mi{T1}),15,180) \in \mi{TV}_{\mi{ON}_1}$ is a possible token visit in $\mi{ON}_1$ shown in \autoref{fig:replay}.
It represents that token $(\mi{p3},\mi{T1}) \in Q_{\mi{ON}_1}$ is produced in place \textit{p3} at $15$ and consumed at $180$.

Given an OCEL, a replay function produces event occurrences and token visits of an OCPN, connecting events in the log to the OCPN.

\begin{definition}[Replay]
Let $\mi{ON}$ be an object-centric Petri net.
A replay function $\mi{replay}_{\mi{ON}} \in \univ{L} \rightarrow \pow(O_{\mi{ON}}) \times \pow(\mi{V}_{\mi{ON}})$ maps an event log to a set of event occurrences and a set of token visits.
\end{definition}

\autoref{fig:replay}(b) shows the result of replaying the events in $L_1$ shown in \autoref{fig:replay}(a) on model $\mi{ON_1}$ depicted in \autoref{fig:replay}(a). 
The dark gray boxes represent event occurrences $O_1$ and the light gray boxes represent token visits $V_1$, where $\mi{replay}_{\mi{ON_1}}(L_1)\allowbreak{=}(O_1,V_1)$.
For instance, replaying event \textit{e1} and \textit{e4} in $L_1$ produces event occurrences, $(\mi{t1},\mi{e1})$ and $(\mi{t3},\mi{e4})$, respectively, and token visit $((\mi{p3},\mi{T1}),15,180)$ where $15$ is the time when \textit{e1} completes and $180$ is the time when \textit{e4} starts.


In this work, we instantiate the replay function based on the token-based replay approach described in~\cite{DBLP:journals/topnoc/BertiA21}.
We first flatten an OCEL to a traditional event log and project an OCPN to an accepting Petri net for each object type.
Next, we apply the token-based replay for each log and Petri net, as introduced in~\cite{DBLP:journals/fuin/AalstB20}.
The replay function needs to be instantiated to ignore non-fitting events to deal with logs with non-perfect fitness.
To simplify matters, we assume the flattened logs perfectly fit the projected Petri nets (i.e., no missing or remaining tokens).

\subsection{Measuring Object-Centric Performance Measures}
We compute object-centric performance measures per event occurrence.
For instance, we compute \emph{synchronization}, \emph{pooling}, \emph{lagging}, and \emph{waiting} time of $(\mi{t3},\mi{e4})$ that analyzes an event of \textit{conduct test}.
For meaningful insights, we may aggregate all waiting time measures of \textit{conduct test} events into the average, median, maximum, or minimum waiting time of \textit{conduct test}.

To this end, we first relate an event occurrence to the token visits 1) associated with the event occurrence's transition and 2) involving the objects linked to the event occurrence's event.

\begin{definition}[Relating An Event Occurrence to Token Visits]
Let $L$ be an object-centric event log and $\mi{ON}$ an object-centric Petri net.
Let $\mi{eo}{=}(t,e) \in O$ be an event occurrence.
$OI(\mi{eo}){=}\bigcup_{\mi{ot} \in \mi{dom}(\pi_{omap}(e))} \pi_{omap}(e)\allowbreak(\mi{ot})$ denotes the set of objects related to the event occurrence.
$rel_{\mi{ON}} \in O_{\mi{ON}} \times \pow(V_{\mi{ON}}) \rightarrow \pow(V_{\mi{ON}})$ is a function mapping an event occurrence and a set of token visits to the set of the token visits related to the event occurrence, s.t., for any $eo \in O_{\mi{ON}}$ and $V \subseteq V_{\mi{ON}}$, $rel_{\mi{ON}}(eo,V){=}\bigcup_{oi \in OI(\mi{eo})} \allowbreak \mi{argmax}_{tv \in \{tv' \in V \mid \pi_{p}(tv') \in \pre t \land \pi_{oi}(tv')=oi\}} \pi_{bt}(tv)$.
\end{definition}

\begin{figure}[!htb]
    \centering
    \includegraphics[width=1\linewidth]{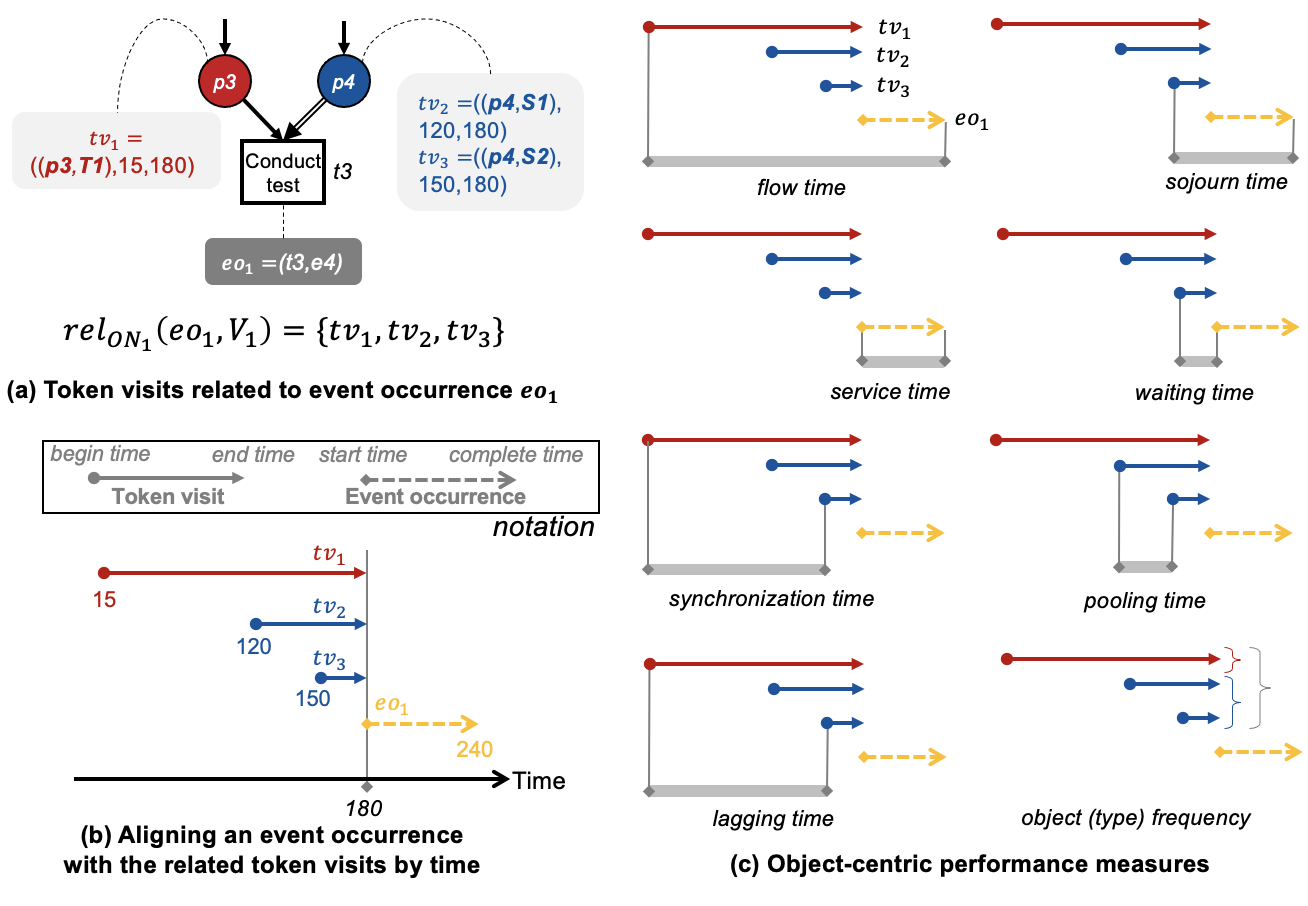}
    \caption{An example of corresponding token visits of an event occurrence and object-centric performance measures of the event occurrence}
    \label{fig:measure}
\end{figure}

\autoref{fig:measure}(a) shows the token visits related to $\mi{eo}_1{=}(\mi{t3},\mi{e4})$.
$rel_{\mi{ON}_1}({\mi{eo}_1,V_1}){=}\{tv_1{=}((\allowbreak \mi{p3} ,\mi{T1}),15,180), tv_2{=}((\mi{p4},\mi{S1}),120,180), tv_3{=}((\mi{p4},\mi{S2}),150,180)\}$ since $\mi{p3},\mi{p4} \allowbreak \in \pre t3$, $\{\mi{T1},\mi{S1},\allowbreak \mi{S2}\} \subseteq OI(\mi{eo}_1)$, and each token visit is with the latest begin time among other token visits of the corresponding object, e.g., $tv_1$ is the latest token visit of $\mi{T1}$.

A measurement function computes a performance measure of an event occurrence by using the related token visits.


\begin{definition}[Measurement]
Let $\mi{ON}$ be an object-centric Petri net.
$\mi{measure} \allowbreak \in O_{\mi{ON}} \times \pow(\mi{V}_{\mi{ON}}) \rightarrow \mathbb{R}$ is a function mapping an event occurrence and its related token visits to a performance value.
$\univ{m}$ denotes the set of all such functions.
\end{definition}

In this paper, we introduce seven measurement functions
to compute object-centric performance measures as shown in \autoref{fig:measure}(c).
With $L$ an OCEL, $\mi{ON}$ an OCPN, and $(O,V){=}\mi{replay}_{\mi{ON}}(L)$,
we introduce the functions with formal definitions and examples as below:

\begin{itemize}
    \item $\mi{flow} \in \univ{m}$ computes \emph{flow time}, i.e., the time difference between the completion of the event and the earliest token visit related to the event. 
    Formally, for any $eo{=}(t,e) \in O$, $\mi{flow}(eo,V)\allowbreak {=} \pi_{ct}(e) - \mi{min}(T)$ with $T{=}\{\pi_{bt}(tv) \mid tv \in rel_{\mi{ON}}(eo,V) \}$.
    In \autoref{fig:measure}(c), the flow time of $eo_1$ is the time difference between the completion of the event, i.e., the completion time of \textit{e4} ($240$), and the earliest token visit related to the event, i.e., the begin time of $tv_1$ ($15$).
    Note that flow time is equal to the sum of \emph{synchronization} time and \emph{sojourn time}.
    
    \item $\mi{sojourn} \in \univ{m}$ computes \emph{sojourn time}, i.e., the time difference between the completion of the event and the latest token visit related to the event.
    Formally, for any $eo{=}(t,e) \in O$, $\mi{sojourn}(eo,\allowbreak V){=} \pi_{ct}(e) - \mi{max}(T)$ with $T{=}\{\pi_{bt}(tv) \allowbreak \mid tv \in rel_{\mi{ON}}(\mi{eo},V) \}$.
    In \autoref{fig:measure}(c), the sojourn time of $eo_1$ is the time difference between the completion of the event, i.e., the completion time of \textit{e4} ($240$), and the latest token visit related to the event, i.e., the begin time of $tk_3$ ($150$).
    Note that sojourn time is equal to the sum of \emph{waiting time} and \emph{service time}.
    
    \item $\mi{wait} \in \univ{m}$ computes \emph{waiting time}, i.e., the time difference between the start of the event and the latest token visit related to the event. 
    Formally, for any $eo{=}(t,e) \in O$, $\mi{wait}(eo,V){=} \pi_{st}\allowbreak(e) - \mi{max}(T)$ with $T{=}\{\pi_{bt}(tv) \mid tv \in rel_{\mi{ON}}(\mi{eo},V) \}$.
    In \autoref{fig:measure}(c), the waiting time of $eo_1$ is the time difference between its start, i.e., the start time of \textit{e4} ($180$), and the latest token visit, i.e., the begin time of $tk_3$ ($150$).
    
    \item $\mi{service} \in \univ{m}$ computes \emph{service time}, i.e., the time difference between the completion of the event and the start of the event.
    Formally, for any $eo{=}(t,e) \in O$, $\mi{service}(eo,V)\allowbreak {=} \pi_{ct}(e) - \pi_{st}(e)$.
    In \autoref{fig:measure}(c), the service time of $eo_1$ is the time difference between the completion of the event, i.e., the completion time of \textit{e4} ($240$), and the start of the event, i.e., the start time of \textit{e4} ($180$).
    
    \item $\mi{sync} \in \univ{m}$ computes \emph{synchronization time}, i.e., the time difference between the latest token visit and the earliest token visit related to the event.
    Formally, for any $eo{=}(t,e) \in O$, $\mi{sync}(eo,V){=} \mi{max}(T)-\mi{min}(T)$ with $T{=}\{\pi_{bt}(tv) \mid tv \in rel_{\mi{ON}}(\mi{eo},V) \}$.
    In \autoref{fig:measure}(c), the synchronization time of $eo_1$ is the time difference between the latest token visit, i.e., the begin time of $tv_3$ ($150$), and the earliest token visit, i.e., the begin time of $tv_1$ ($15$).
    Note that the synchronization time consists of \emph{pooling time} and \emph{lagging time}.
    
    \item $\mi{pool}_{ot} \in \univ{m}$ computes \emph{pooling time} w.r.t. object type $\mi{ot}$, i.e., the time difference between the latest token visit of $\mi{ot}$ and the earliest token visit of $\mi{ot}$ related to the event. 
    Formally, for any $eo{=}(t,e) \in O$, $\mi{pool}_{\mi{ot}}(eo,V){=} \mi{max}(T)-\mi{min}(T)$ with $T{=}\{\pi_{bt}(tv) \mid tv \in rel_{\mi{ON}}(\mi{eo},V) \allowbreak \wedge \mi{type}(\pi_{oi}(tv)){=}\mi{ot} \}$.
    In \autoref{fig:measure}(c), the pooling time of $eo_1$ w.r.t. \textit{sample} is the time difference between the latest token visit of \textit{sample}, i.e., the begin time of $tv_3$ ($150$), and the earliest token visit of \textit{sample}, i.e., the begin time of $tv_2$ ($120$).
    Note that the pooling time can be the same as the synchronization time.
    
    \item $\mi{lag}_{ot} \in \univ{m}$ computes \emph{lagging time} w.r.t. object type $\mi{ot}$, i.e., the time difference between the latest token visit of $\mi{ot}$ and the earliest token visit of other object types related to the event. 
    Formally, for any $eo{{=}}(t,e) \in O$, $\mi{lag}_{\mi{ot}}(eo,V){=} \mi{max}(T')-\mi{min}(T)$ with $T{=}\{\pi_{bt}(tv) \mid tv \in rel_{\mi{ON}}(\mi{eo},V) \}$ and $T'{=}\{\pi_{bt}(tv) \mid tv \in rel_{\mi{ON}}(\mi{eo},V) \wedge \mi{type}(\pi_{oi}(tv)){\neq}\mi{ot}\}$ if $\mi{max}(T')>\mi{min}(T)$. $0$ otherwise.
    In \autoref{fig:measure}(c), the lagging time of $eo_1$ w.r.t. \textit{sample} is the time difference between the latest token visit of \textit{samples}, i.e., the begin time of $tv_3$ ($150$), and the earliest token visit of any object types, i.e., the begin time of $tv_1$ ($15$).
    Note that, in some cases, the lagging time is the same as the synchronization time.
\end{itemize}

Non-temporal performance measures are trivial to compute given object-centric event data, but still provide valuable insights. They include \emph{object frequency}, i.e., the number of objects involved with the event, and \emph{object type frequency}, i.e., the number of object types involved with the event.
In \autoref{fig:measure}(c), the object frequency of \textit{e4} is $3$ including \textit{T1}, \textit{S1}, and \textit{S2} and the object type frequency of \textit{e4} is $2$ including \textit{Test} and \textit{Sample}.

%% file: Sections/5-Evaluation.tex
\section{Evaluation}\label{sec:evaluation}
In this section, we present the implementation of the proposed approach and evaluate the effectiveness of the approach by applying it to a real-life event log.

\subsection{Implementation}
The approach discussed in \autoref{sec:approach} has been fully implemented as a web application\footnote{A demo video, sources, and manuals are available at \url{https://github.com/gyunamister/OPerA}} with a dedicated user interface. 
We containerize it as a Docker container, structuring functional components into a coherent set of microservices.
The following functions are supported:

\begin{itemize}
    \item Importing object-centric event logs in different formats including OCEL JSON, OCEL XML, and CSV.
    \item Discovering OCPNs based on the general approach presented in~\cite{DBLP:journals/fuin/AalstB20} with Inductive Miner Directly-Follows process discovery algorithm~\cite{DBLP:journals/sosym/LeemansFA18}.
    \item Replaying tokens with timestamps on OCPNs based on token-based replay approach suggested in~\cite{DBLP:journals/topnoc/BertiA21}.
    \item Computing object-centric performance measures based on the replay results, i.e., event occurrences and token visits.
    \item Visualizing OCPNs with the object-centric performance measure.
\end{itemize}

\begin{figure}
    \includegraphics[width=1\linewidth]{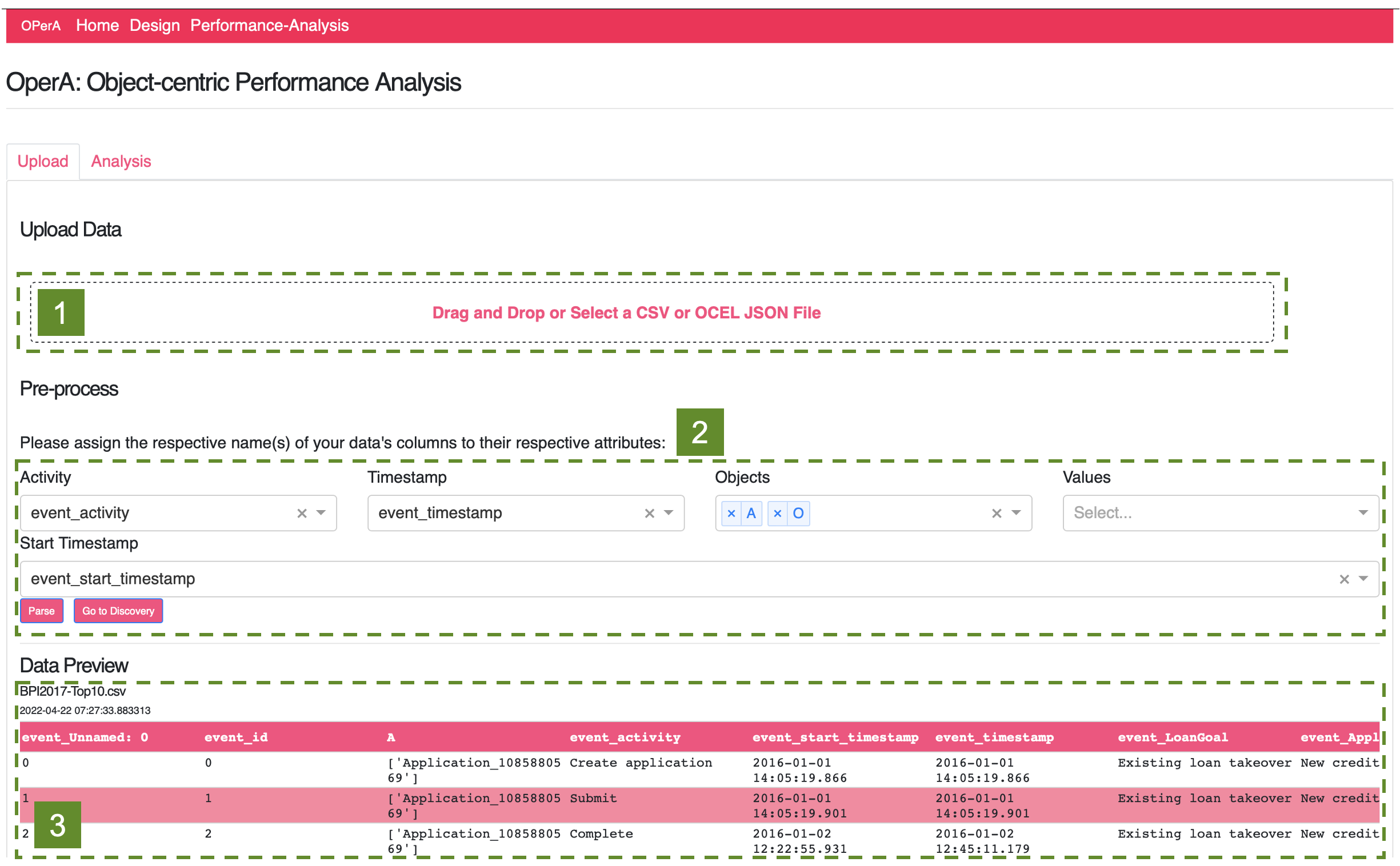}
\caption{
A screenshot of the application: Importing Object-Centric Event Logs (OCELs). 
\raisebox{.3pt}{\boxed{\raisebox{-.6pt} {1}}} Importing OCELs in OCEL JSON, OCEL XML, and CSV formats.
\raisebox{.3pt}{\boxed{\raisebox{-.6pt} {2}}} Preprocessing OCELs.
\raisebox{.3pt}{\boxed{\raisebox{-.6pt} {3}}} Displaying OCELs.
}
\label{fig:application-1}
\end{figure}

\begin{sidewaysfigure}
    \includegraphics[width=1\linewidth]{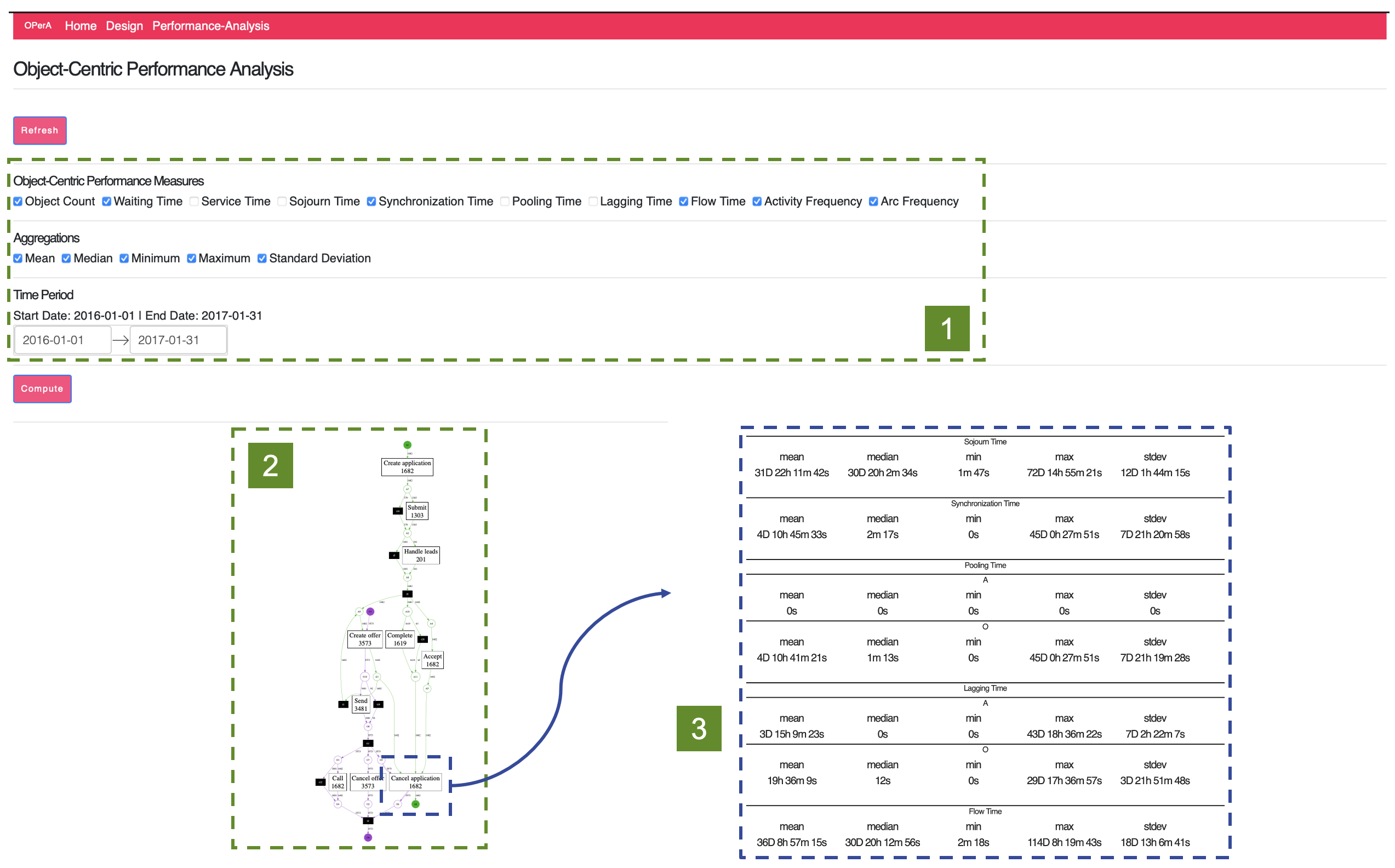}
\caption{
A screenshot of the application: Analyzing and visualizing object-centric performance measures. 
\raisebox{.3pt}{\boxed{\raisebox{-.6pt} {1}}} Selecting object-centric performance measures, aggregations, and a time period to analyze.
\raisebox{.3pt}{\boxed{\raisebox{-.6pt} {2}}} An object-centric Petri net visualizing the computed performance measures.
\raisebox{.3pt}{\boxed{\raisebox{-.6pt} {3}}} Visualizing the detailed performance measures of a selected activity from the model.
}
\label{fig:application-2}
\end{sidewaysfigure}

\subsection{Case Study: Loan Application Process}
Using the implementation, we conduct a case study on a real-life loan application process of a Dutch Financial Institute\footnote{\url{doi.org/10.4121/uuid:3926db30-f712-4394-aebc-75976070e91f}}.
Two object types exist in the process: \textit{application} and \textit{offer}. 
An application can have one or more offers.
First, a customer creates an application by visiting the bank or using an online system. In the former case, \textit{submit} activity is skipped.
After the completion and acceptance of the application, the bank offers loans to the customer by sending the offer to the customer and making a call.
An offer is either accepted or canceled.

In this case study, we focus on the offers canceled due to various reasons.
We filter infrequent behaviors by selecting the ten most frequent types of process executions.
Moreover, we remove redundant activities, e.g., status updates such as \textit{Completed} after \textit{Complete application}.
The resulting event log, available at the Github repository, contains $20,478$ events by $1,682$ applications and $3,573$ offers.

\begin{figure}
    \includegraphics[width=1\linewidth]{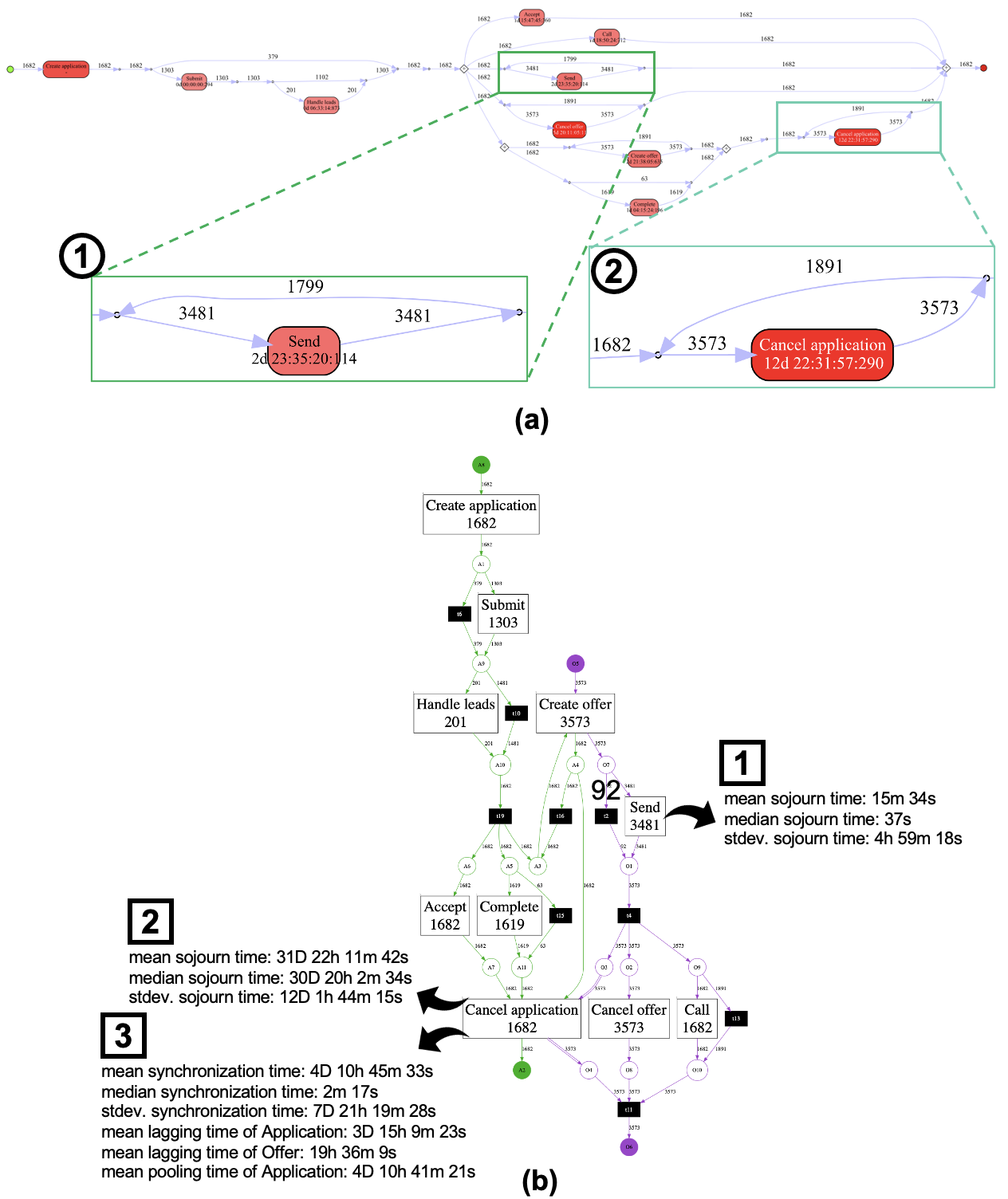}
\caption{(a) Performance analysis results based on \textit{Inductive Visual Miner} in \textit{ProM} framework and (b) Performance analysis results based on our proposed approach. We compare \raisebox{.3pt}{\textcircled{\raisebox{-.6pt} {1}}}, \raisebox{.3pt}{\textcircled{\raisebox{-.6pt} {2}}}, and \raisebox{.3pt}{\textcircled{\raisebox{-.6pt} {3}}} with \raisebox{.3pt}{\boxed{\raisebox{-.6pt} {1}}}, \raisebox{.3pt}{\boxed{\raisebox{-.6pt} {2}}}, and \raisebox{.3pt}{\boxed{\raisebox{-.6pt} {3}}}, respectively.
\raisebox{.3pt}{\boxed{\raisebox{-.6pt} {4}}} shows the result on newly-introduced performance measures.}
\label{fig:experiment}
\end{figure}

First, we compare our approach to a traditional technique for performance analysis based on alignments~\cite{6eac4d541d66416180c36e95696ea87f}.
To apply the traditional technique, we first flatten the log using the application and offer as a case notion.
\autoref{fig:experiment}(a) shows the performance analysis results from \textit{Inductive Visual Miner} in \textit{ProM} framework\footnote{\url{https://www.promtools.org}}.
As shown in \raisebox{.5pt}{\textcircled{\raisebox{-.9pt} {1}}}, $1,799$ applications repeat activity \textit{Send}. 
In reality, as shown in \raisebox{.5pt}{\boxed{\raisebox{-.9pt} {1}}}, no repetition occurs while the activity is conducted once for each offer except $92$ offers skipping it.
Furthermore, the average sojourn time for the activity is computed as around $2$ days and $23$ hours, whereas, in reality, it is around $15$ minutes as shown in \raisebox{.5pt}{\boxed{\raisebox{-.9pt} {1}}}.

Furthermore, \raisebox{.5pt}{\textcircled{\raisebox{-.9pt} {2}}} shows that activity \textit{Cancel application} is repeated $1891$ times, but it occurs, in reality, $1,682$ times for each application, as depicted in \raisebox{.5pt}{\boxed{\raisebox{-.9pt} {2}}}.
In addition, the average sojourn time for the activity is measured as around $12$ days and $22$ hours, but in fact, it is around $31$ days and $22$ hours, as shown in \raisebox{.5pt}{\boxed{\raisebox{-.9pt} {2}}}.

Next, we analyze the newly-introduced object-centric performance measures, including synchronization, lagging, and pooling time.
As described in \raisebox{.5pt}{\boxed{\raisebox{-.9pt} {3}}}, the average synchronization time of activity \textit{Cancel application} is around $4$ days and $11$ hours.

Moreover, the average lagging time of \textit{applications} is $3$ days and $15$ hours and the lagging time of \textit{offers} is $19$ hours, i.e., \textit{offers} are more severely lagging \textit{applications}.
Furthermore, the pooling time of \textit{offers} is almost the same as the synchronization time, indicating that the application is ready to be cancelled almost at the same time as the first offer, and the second offer is ready in around $4$ days and $11$ hours.

%% file: Sections/6-Conclusion.tex
\section{Conclusion}\label{sec:conclusion}
In this paper, we proposed an approach to object-centric performance analysis, supporting the correct computation of existing performance measures and the derivation of new performance measures considering the interaction between objects.
To that end, we first replay OCELs on OCPNs to couple events to process models, producing event occurrences and token visits.
Next, we measure object-centric performance metrics per event occurrence by using the corresponding token visits of the event occurrence.
We have implemented the approach as a web application and conducted a case study using a real-life loan application process of a financial institute.

The proposed approach has several limitations.
First, our approach relies on the quality of the discovered process model.
Discovering process models that can be easily interpreted and comprehensively reflect the reality is a remaining challenge.
Second, non-conforming behavior in event data w.r.t. a process model can lead to misleading insights.
If \textit{Transfer samples} is missing for a sample in an event log, although a process model describes that it always occurs for samples, the performance measure of \textit{Clear sample} w.r.t. the sample will be computed based on the wrong timestamps from \textit{Conduct Test}.
In the implementation, we use process discovery techniques that guarantee the discovery of a perfectly-fitting process model and remove the issue of non-conforming behavior.
As future work, we plan to extend the approach to support reliable performance analysis of non-conforming event logs.
Moreover, we plan to develop an approach to object-centric performance analysis based on event data independently from process models.
Another direction of future work is to define and compute more interesting performance metrics that consider the interaction between objects.
